\documentclass[conference]{IEEEtran}
\usepackage[numbers,sort&compress]{natbib}
\usepackage{amsmath,amssymb,amsfonts}
\usepackage{algorithmic}
\usepackage{graphicx}
\usepackage{textcomp}
\usepackage{xcolor}
\def\BibTeX{{\rm B\kern-.05em{\sc i\kern-.025em b}\kern-.08em
    T\kern-.1667em\lower.7ex\hbox{E}\kern-.125emX}}

\usepackage{algorithm}
\usepackage{algorithmic}
\usepackage{orcidlink}

\begin{document}

\title{Can Local Representation Alignment RNNs Solve Temporal Tasks?}

\author{\IEEEauthorblockN{1\textsuperscript{st} Nikolay Manchev*}
\IEEEauthorblockA{\textit{Department of Informatics} \\
\textit{King's College London}\\
London, UK \\
\href{https://orcid.org/0000-0003-2987-9234}{ORCID: 0000-0003-2987-9234}}
*Corresponding author
~\\
\and
\IEEEauthorblockN{2\textsuperscript{nd} Luis C. Garcia-Peraza-Herrera}
\IEEEauthorblockA{\textit{Department of Informatics} \\
\textit{King's College London}\\
London, UK \\
\href{https://orcid.org/0000-0002-7362-0353}{ORCID: 0000-0002-7362-0353}
}
~\\
}

\maketitle

\begin{abstract}
Recurrent Neural Networks (RNNs) are commonly used for 
real-time processing, streaming data, and cases where the amount of training samples is limited. Backpropagation Through Time (BPTT) is the predominant algorithm for training RNNs; however, it is frequently criticized for being prone to exploding and vanishing gradients and being biologically implausible. 
In this paper, we present and evaluate a target propagation-based method for RNNs, which uses local updates and seeks to reduce the said instabilities. 
Having stable RNN models increases their practical use in a wide range of fields such as natural language processing, time-series forecasting, anomaly detection, control systems, and robotics. 

The proposed solution uses local representation alignment (LRA). 
We thoroughly analyze the performance of this method, experiment with normalization and different local error functions, and invalidate certain assumptions about the behavior of this type of learning. Namely, we demonstrate that despite the decomposition of the network into sub-graphs, the model still suffers from vanishing gradients. We also show that gradient clipping as proposed in LRA has little to no effect on network performance. This results in an LRA RNN model that is very difficult to train due to vanishing gradients.
We address this by introducing gradient regularization in the direction of the update and demonstrate that this modification promotes gradient flow and meaningfully impacts convergence. 
We compare and discuss the performance of the algorithm, and we show that the regularized LRA RNN considerably outperforms the unregularized version on three landmark synthetic tasks.
\end{abstract} 

\section{Introduction}

RNNs were conceptualized in the work of Jordan~\cite{jordan1986attractor}, Elman~\cite{elman1990finding}, Werbos~\cite{werbos1988generalization} and others, and have shown promising results on a wide range of sequential learning problems such as text generation \cite{shini2021recurrent, boopathi2023deep, parmar2024novel}, machine translation \cite{xiao2020research, mallick2021context, wang2022progress}, unsupervised feature extraction and action recognition using video data \cite{sharma2021video, ullah2021efficient}, drug molecule generation \cite{zou2023generation}, time series prediction \cite{HEWAMALAGE2021388, wang2022ngcu, dudukcu2023temporal}, and speech recognition and synthesis \cite{saon2021advancing, oruh2022long, khanam2022text}.

At the same time, RNNs have long been criticized for being biologically implausible \cite{whittington2019theories, pulvermuller2021biological, schmidgall2024brain}. 
This judgment stems mostly from the backpropagation technique, which is typically employed to train RNNs. 
According to Crick~\cite{Crick1989}, backpropagation-trained neural networks are biologically unrealistic ``in almost every respect''. Shervani-Tabar \& Rosenbaum~\cite{shervani2023meta} point out that backpropagation's ``relationship to synaptic plasticity in the brain is unknown'', and Capone \& Paolucci~\cite{capone2024towards} state that algorithms that rely on non-local learning rules are ``computationally expensive and biologically implausible''.
Additional challenges involve the requirement that each neuron produces dual signals (output and error), the credit assignment problem, and the weight transport mechanisms that require knowledge of all synaptic weights in the forward path \cite{Minsky1961,Hinton1984,Lillicrap2016}.

RNNs are also notoriously difficult to train \cite{hochreiter2001gradient}, as chain rule-based backpropagation ultimately leads to gradient instabilities, exacerbated by increased depth of the network~\cite{pmlr-v28-pascanu13, mikhaeil2022difficulty}. 
Such instabilities could be mitigated by careful initialization \cite{10.5555/3305890.3306050, 10068489} or by employing target-based optimization like Target Propagation Through Time (TPTT) \cite{JMLR:v21:18-141}, which avoids backpropagation altogether. 

\textbf{Contributions.} We propose a novel LRA-based method for training RNNs and solving tasks with temporal dependencies.
This algorithm is deemed more biologically plausible as it performs local updates and does not rely on global error signals. This is valuable because linking RNNs to biological networks can help improve existing recurrent models with insights from neuroscience. In contrast, biologically plausible RNNs can drive a better understanding of cognition \cite{cohen2022recent}.
Our method further improves LRA training by introducing a regularization scheme that increases its original performance and makes it less susceptible to vanishing gradients. This facilitates the training of deeper networks and mitigates gradient instabilities, thus increasing the overall effectiveness of LRA networks. In addition, we introduce an extra hyperparameter to control the target update step. This can eliminate the computationally expensive inner loop of LRA and make it converge faster.
We compare this improved version with the state of the art target propagation algorithm for RNNs (TPTT) and provide evaluation results. This includes testing LRA beyond the trivial depth of eight layers. We show that for the random permutation task, the regularized LRA outperforms BPTT. We also show evidence that, despite the sub-graph decomposition, the vanilla LRA algorithm is prone to vanishing gradients, which contradicts previous claims in \cite{ororbia2018conducting}. 
\section{Related Work}
Target propagation is an alternative learning approach to backpropagation that addresses some of its key challenges, such as reliance on the chain rule and the need for global error signals. The main mechanism of target propagation, the idea to associate targets instead of loss gradients with the feedforward activations, was initially proposed by LeCun~\cite{266b3d0e70ff48af8a24ba238f7e7222, yann1987modeles}. Lee et al.~\cite{lee2015difference} propose a variant of target propagation called Difference Target Propagation (DTP), designed to address the difficulty of generating accurate targets for intermediate layers. Their method introduces a difference correction term, ensuring that layer-wise targets are more consistent with the global objective. Manchev \& Spratling~\cite{JMLR:v21:18-141} present a method for using DTP in RNNs, which significantly outperforms backpropagation in terms of model performance and convergence speed. The authors also demonstrate that TPTT can ``unlock'' RNNs by removing the dependence on precisely clocked forward and backward phases, thus paving the way toward training parallelization. Continuing the line of investigation of target propagation in RNNs, Manchev \& Spratling~\cite{https://doi.org/10.1111/coin.12691} design two architectures that enable RNNs to solve multi-model problems with temporal components. However, the exact computation of targets and assumptions about weight symmetry question the compatibility of TPTT with neural processes.

The concept of feedback alignment was introduced by Lillicrap et al.~\cite{Lillicrap2016}, who show that random feedback weights can be used instead of symmetric weights for propagating error signals. Expanding on this, N{\o}kland~\cite{nokland2016direct} proposes a modification named Direct Feedback Alignment (DFA), which feeds error signals directly from the output layer to the hidden layer through fixed random connections. N{\o}kland demonstrates that DFA achieves model performance closely matching backpropagation, while providing a step towards biologically plausible machine learning. However, numerous studies show that DFA consistently fails to match backpropagation in convolutional neural networks \cite{Bartunov2018AssessingTS, launay2019principled, han2019direct}. Refinetti et al.~\cite{refinetti2021align} develop a theoretical framework for understanding feedback alignment. Their work provides insight into how the data structure impacts alignment and explains the failures observed when training convolutional networks with feedback alignment.

LRA is a family of algorithms closely related to feedback alignment and target propagation. LRA was introduced by Ororbia et al.~\cite{ororbia2018conducting} in a standard feedforward neural network. It can be adapted for RNNs by means of unrolling \cite{robinson:utility, 58337}, but its use in RNNs has not been formally developed or studied.
LRA variants employ feedback matrices to update targets rather than weights. LRA also uses a layer-wise target assignment strategy reminiscent of target propagation. Unlike other target propagation algorithms like TPTT, where the local targets are set in a way that reduces the global loss, LRA sets targets in a way that minimizes the loss only in the next downstream layer. The expectation is that this creates a cascade effect, leading to the final layer where minimization of the local loss is equivalent to minimizing the global network loss.
LRA is considered more biologically plausible, as it relies only on error signals in close proximity to the trained neuron. This approach is closer to Hebbian learning \cite{hebb-organization-of-behavior-1949}, which aligns better with observed patterns in the synaptic plasticity of the brain and has been used as a basis for various mechanisms that model neural activity \cite{abbott2000synaptic}.
Ororbia et al.~\cite{ororbia2018conducting} claim that LRA-based algorithms completely avoid gradient instability problems, making them especially suitable for training deep neural networks. This is based on the assumption that because LRA treats the network as a collection of shallow sub-graphs, vanishing gradients do not manifest themselves (i.e. all optimization is constrained to learning a collection of two-layer networks). However, this is not substantiated by empirical evidence as the deepest network LRA is tested with is relatively shallow --- only 8 hidden layers. 
Three different variants of LRA have been suggested: \texttt{LRA-diff} (backpropagation adaptation given in \cite{ororbia2018conducting}), \texttt{LRA-fdbk} (using a feedback alignment mechanism, Ibid.), and \texttt{LRA-E} (error-driven LRA, proposed in \cite{Ororbia_Mali_2019}). Here we focus on \texttt{LRA-diff}, as it is the closest to BPTT in the manner it calculates targets and handles weight updates. This enables us to evaluate its performance using TPTT and BPTT as baselines. 
\section{Local Representation Alignment in Recurrent Neural Networks}

In this section, we present \texttt{RNN-LRA-diff}, a novel method for training RNNs using local representation alignment. RNNs process sequential data by maintaining a hidden state that captures information about the sequence. Let $\{\boldsymbol{x}_1, \boldsymbol{x}_2, \dots, \boldsymbol{x}_T\}$ be an input sequence where $\boldsymbol{x}_t \in \mathbb{R}^n$. At each time step $t \in [1,t_{max}] $, the hidden state $\boldsymbol{h}_t \in \mathbb{R}^m$ is updated based on the current input $\boldsymbol{x}_t$ and the previous hidden state $\boldsymbol{h}_{t-1}$:

\begin{align}\label{eq:forward-rnn}
\begin{split}
\boldsymbol{h}_t &= \boldsymbol{z}_{t-1} \cdot \boldsymbol{W}_{hh} + \boldsymbol{x}_t \cdot \boldsymbol{W}_{xh} + \boldsymbol{b}_h \\
\boldsymbol{z}_t &= \sigma(\boldsymbol{h}_t)
\end{split}
\end{align}

\noindent where $\boldsymbol{W}_{xh} \in \mathbb{R}^{m\times n}$ is a weight matrix for the input, $\boldsymbol{W}_{hh} \in \mathbb{R}^{m\times m}$ is a transition state matrix, $\boldsymbol{b}_h \in \mathbb{R}^m$ is a bias vector for the hidden state, and $\sigma$ is a nonlinear element-wise activation function. The starting values of these parameters can be assigned using various initialization schemes such as Xavier~\cite{pmlr-v9-glorot10a} or orthogonal initialization \cite{pmlr-v48-henaff16, pmlr-v48-arjovsky16}. The starting hidden state $\boldsymbol{h}_0$ is typically initialized with zeros. At time step $t_{max}$, the network uses an output mapping matrix $\boldsymbol{W}_{hy} \in \mathbb{R}^{k\times m}$ and a bias vector $\boldsymbol{b}_y \in \mathbb{R}^k$ to calculate its prediction $\hat{\boldsymbol{y}} \in \mathbb{R}^k$ as

\begin{align}
\hat{\boldsymbol{y}} = z_{t_{max}} \cdot \boldsymbol{W}_{hy} + \boldsymbol{b}_y
\end{align}

\noindent optionally applying the normalized exponential function when training on classification problems. A global loss $\mathcal{E} (\hat{\boldsymbol{y}},\boldsymbol{y})$ can then be calculated using a problem-appropriate error function.

\begin{figure}[ht]
\begin{center}
\centerline{\includegraphics[width=\columnwidth]{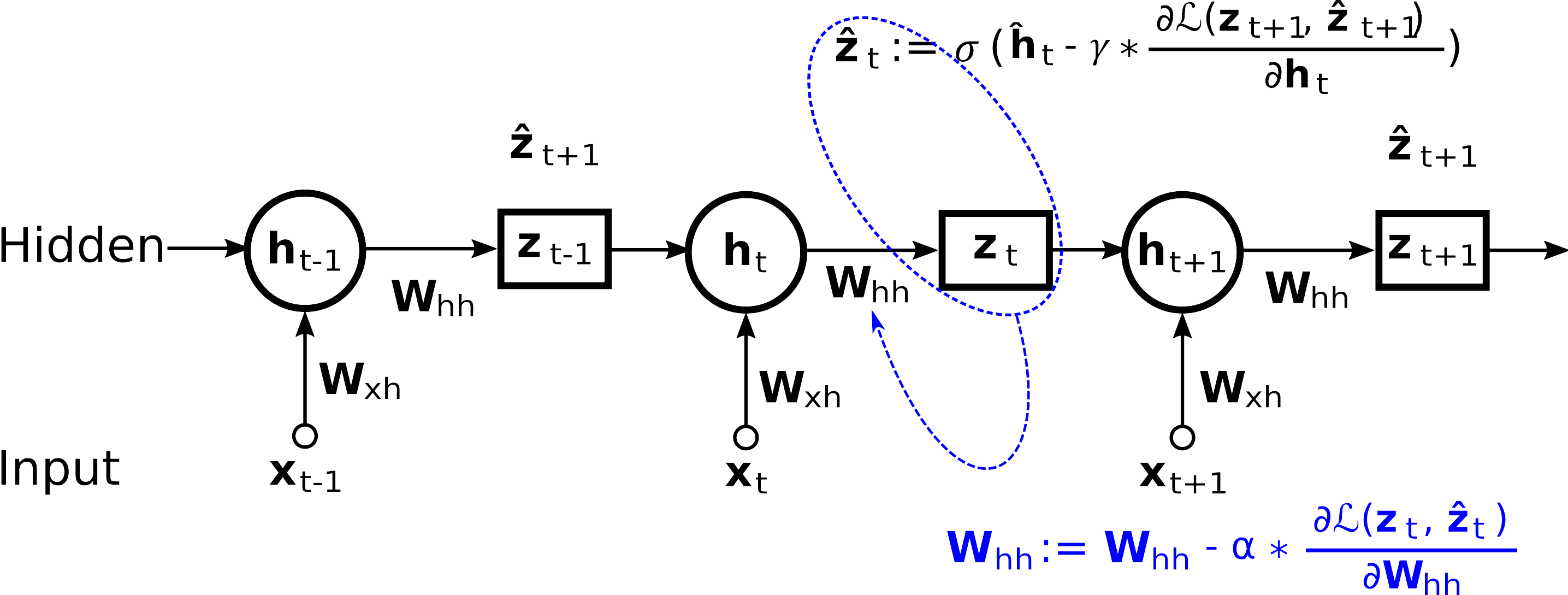}}
\caption{LRA-diff in recurrent neural networks. The transition and input-mapping matrices are updated to minimize $\mathcal{L} (\boldsymbol{z}_t, \hat{\boldsymbol{z}}_t)$. \textbf{Best viewed online.}}
\label{fig:LRA-RNN}
\end{center}
\end{figure}

We will now discuss how \texttt{LRA-diff} can be applied to a recurrent neural network. Consider a simple RNN unrolled over $t_{max}$ time steps. With standard backpropagation through time, the updates of the upstream layers are estimated using 

\begin{equation}
	\frac{\partial \mathcal{E}}{\partial \textbf{W}_{hh}} = \sum\limits_{t = 1}^{t_{\text{max}}} \frac{\partial \mathcal{E}}{\partial \boldsymbol{y}} \frac{\partial \boldsymbol{y}}{\partial \textbf{h}_{t_{\text{max}}}} \frac{\partial \textbf{h}_{t_{\text{max}}}}{\partial \textbf{h}_t} \frac{\partial \textbf{h}_t}{\partial \textbf{W}_{hh}}
\end{equation}

\begin{algorithm}
\caption{The LRA-diff algorithm applied to RNNs}\label{alg:lra-diff}
\textbf{Input:} Dataset $\{\textbf{x}, y\}$, transition matrix $\boldsymbol{W}_{hh}$, input mapping matrix $\boldsymbol{W}_{xh}$, output mapping matrix $\boldsymbol{W}_{hy}$, bias terms $\boldsymbol{b}_h$ and $\boldsymbol{b}_y$, nonlinear activation function $\sigma(\cdot)$, number of steps $K$, target adjustment step $\gamma$, normalization constants $c_0$ and $c_1$,  global loss function $\mathcal{E(\cdot)}$, local loss function $\mathcal{L(\cdot)}$\\
\textbf{Output:} Gradient updates $\{\nabla \boldsymbol{W}_{xh}$, $\nabla \boldsymbol{W}_{hh}$, $\nabla \boldsymbol{W}_{hy}$, $\nabla \boldsymbol{b}_{h}$, $\nabla \boldsymbol{b}_{y}\}$\\
    \begin{algorithmic}[1]
        \STATE \textbf{function} \texttt{GradUpdates}($\{\textbf{x}, y\}$, $\boldsymbol{W}_{xh}$, $\boldsymbol{W}_{hh}$, $\boldsymbol{W}_{hy}$, $\boldsymbol{b}_{h}$, $\boldsymbol{b}_{y}$,
             $\sigma(\cdot)$, $K$, $\gamma$, $c_0$, $c_1$, $\mathcal{E(\cdot)}$, $\mathcal{L(\cdot)}$)
        \STATE \hspace{1em} $\boldsymbol{z}_0 \gets \sigma(\boldsymbol{x}_0 \cdot \boldsymbol{W}_{xh} + \boldsymbol{b}_h)$
        \vspace{1mm}
        \STATE \hspace{1em} \textbf{for } {$t \gets 1$ \textbf{ to } $t_{max}$} \textbf{ do} 
        \STATE \hspace{2em} $\boldsymbol{h}_t \gets \boldsymbol{z}_{t-1} \cdot \boldsymbol{W}_{hh} + \boldsymbol{x}_t \cdot \boldsymbol{W}_{xh} + \boldsymbol{b}_h$
        
        \STATE \hspace{2em} $\boldsymbol{z}_t \gets \sigma(\boldsymbol{h}_t)$
        
        \STATE \hspace{1em} $\hat{\boldsymbol{y}} \gets z_{t_{max}} \cdot \boldsymbol{W}_{hy} + \boldsymbol{b}_y$
        \item[]
        \STATE \hspace{1em} $\nabla \boldsymbol{W}_{hy} \gets \frac{\partial\mathcal{E(\hat{\boldsymbol{y}}, \boldsymbol{y})}}{\partial \boldsymbol{W}_{hy}}$, $\nabla \boldsymbol{b}_{y} \gets \frac{\partial\mathcal{E(\hat{\boldsymbol{y}}, \boldsymbol{y})}}{\partial \boldsymbol{b}_{y}}$ 
        \STATE \hspace{1em} $\bar{\boldsymbol{h}}_{t_{max}} \gets \boldsymbol{h}_{t_{max}}$ 
        \STATE \hspace{1em} \textbf{for } {$k \gets 1$ \textbf{ to } $K$} \textbf{ do} 
        \STATE \hspace{2em} $\Delta \boldsymbol{h}_{h_{t_{max}}} \gets \frac{\partial \mathcal{E(\hat{\boldsymbol{y}}, \boldsymbol{y})}}{ \partial \bar{\boldsymbol{h}}_{t_{max}} }$
        \STATE \hspace{2em} $\hat{\boldsymbol{h}}_{t_{max}} \gets \bar{\boldsymbol{h}}_{t_{max}} - \gamma \cdot  \Delta \boldsymbol{h}_{h_{t_{max}}}$
        \STATE \hspace{2em} $\hat{\boldsymbol{z}}_{t_{max}} \gets \sigma(\hat{\boldsymbol{h}}_{t_{max}})$ 
        \STATE \hspace{2em} $\hat{\boldsymbol{y}} \gets z_{t_{max}} \cdot \boldsymbol{W}_{hy} + \boldsymbol{b}_y$
        \item[]
        \STATE \hspace{1em} $t \gets t_{max}$ \STATE \hspace{1em}    \textbf{while }{$t \geq 1$} \textbf{ do }
        \STATE \hspace{2em} $\nabla \boldsymbol{W}^t_{hh} \gets \text{Normalize} \left[\frac{\partial\mathcal{L}(\boldsymbol{z}_t, \hat{\boldsymbol{z}}_t)}{\partial \boldsymbol{W}_{hh}}, c_0\right]$
        \STATE \hspace{2em} $\nabla \boldsymbol{b}^t_{h} \gets \text{Normalize} \left[\frac{\partial \mathcal{L}(\boldsymbol{z}_t, \hat{\boldsymbol{z}}_t)}{\partial \boldsymbol{b}_{h}}, c_0\right]$ \label{lst:normalise1}
        \STATE \hspace{2em} $\nabla \boldsymbol{W}^t_{xh} \gets \text{Normalize} \left[\frac{\partial \mathcal{L}(\boldsymbol{z}_t, \hat{\boldsymbol{z}}_t) }{\partial \boldsymbol{W}_{xh}}, c_0\right]$ \label{lst:normalise2}
        \STATE \hspace{2em} $\bar{\boldsymbol{h}}_{t-1} \gets \boldsymbol{h}_{t-1}$ 
        \STATE \hspace{2em} \textbf{for } {$k \gets 1$ \textbf{ to } $K$} \label{lst:inner-loop} \textbf{ do}
        \STATE \hspace{3em} $\Delta \boldsymbol{h}_{t-1} \gets \text{Normalize} \left[\frac{\partial \mathcal{L}(\boldsymbol{z}_t, \hat{\boldsymbol{z}}_t)}{ \partial \bar{\boldsymbol{h}}_{t-1} }, c_1\right]$ \label{lst:normalise3}
        \STATE \hspace{3em} $\hat{\boldsymbol{h}}_{t-1} \gets \bar{\boldsymbol{h}}_{t-1} - \gamma \cdot  \Delta \boldsymbol{h}_{t-1}$
        \STATE \hspace{3em} $\hat{\boldsymbol{z}}_{t-1} \gets \sigma(\hat{\boldsymbol{h}}_{t-1})$         
        \STATE \hspace{3em} $\hat{\boldsymbol{h}_t} \gets\hat{\boldsymbol{z}}_{t-1} \cdot \boldsymbol{W}_{hh} + \boldsymbol{x}_t \cdot \boldsymbol{W}_{xh} + \boldsymbol{b}_h$
        \STATE \hspace{3em} $\hat{\boldsymbol{z}}_t \gets \sigma(\hat{\boldsymbol{h}}_t)$
        \STATE \hspace{3em} $\bar{\boldsymbol{h}}_{t-1} \gets \hat{\boldsymbol{h}}_{t-1}$ 
        \STATE \hspace{2em} $t \gets t - 1$
        \vspace{1mm}
        \STATE \hspace{1em}  \textbf{return } \{$\sum\limits_t \nabla \boldsymbol{W}^t_{xh}$, $\sum\limits_t \nabla \boldsymbol{W}^t_{hh}$, $\nabla \boldsymbol{W}_{hy}$, \\ \hspace{4.8em} $\sum\limits_t \nabla \boldsymbol{b}^t_{h}$, $\nabla \boldsymbol{b}_{y}$\}
        \item[]
        \STATE \textbf{function} \texttt{Normalize} ({$\Delta$, c}) \label{lst:normalise}   
        \vspace{1mm}
        \STATE \hspace{1em} \textbf{If} {$c>0$ \textbf{and} $||\Delta|| \geq c$} \textbf{then}
        \STATE \hspace{2em} $\Delta \gets \frac{c}{||\Delta||}\Delta$
        \vspace{1mm}
        \STATE \hspace{1em} \textbf{return} $\Delta$
    \end{algorithmic}
\end{algorithm}

Calculating $\frac{\partial \boldsymbol{h}_{t_{max}}}{\partial \boldsymbol{h}_t}$ potentially leads to gradient instabilities for temporally distant events \cite{pmlr-v28-pascanu13}. To combat this, \texttt{LRA-diff} minimizes a local loss $\mathcal{L}(\boldsymbol{z}_t, \hat{\boldsymbol{z}}_t)$. This idea is illustrated in Figure~\ref{fig:LRA-RNN}: the targets $\hat{\boldsymbol{z}}_t$ are chosen by implicitly selecting values for time step $t$ in a way that brings the calculation at time step $t+1$ closer to its respective target. $\hat{\boldsymbol{z}}_t$ is given by

\begin{equation}\label{eq:targets}
    \hat{\boldsymbol{z}}_t = \sigma\left(\hat{\boldsymbol{h}}_t - \gamma * \frac{\partial \mathcal{L} (\boldsymbol{z}_{t+1}, \hat{\boldsymbol{z}}_{t+1})}{\partial \boldsymbol{h}_t}\right)
\end{equation}

Inspired by \texttt{LRA-diff}, we introduce a novel training approach called \texttt{RNN-LRA-diff} that specifically targets RNNs. The detailed steps of our proposed method are shown in Algorithm~\ref{alg:lra-diff}.
The original LRA algorithm uses an inner loop, which refines the targets over $K$ iterations. According to Ororbia et al.~\cite{ororbia2018conducting}, increasing the value of $K$ leads to a ``significant slowdown'' and ``diminishing return'' for $K > 30$. To address this limitation, we introduce a hyperparameter $\gamma$ in \eqref{eq:targets}, used for fine-grained control of the target update step in the inner loop. This additional parameter can be used to eliminate the inner loop (e.g. setting $K=1$) whilst retaining control over the speed of change of the targets. As we focus our attention on the suitability of \texttt{RNN-LRA-diff} for maintaining stable gradient updates and training recurrent networks of greater depth, we also test LRA for much deeper networks than the 8-layer architecture used in \cite{ororbia2018conducting}. 

\subsection{Regularizing vanishing gradients}\label{sec:regularising}

Our experiments show that \texttt{RNN-LRA-diff} performs poorly with the increase of sequence length. To address this, we design a fix for the vanishing gradient problem based on \cite{pmlr-v28-pascanu13}. They present a regularizer

\begin{equation}\label{eq:omega}
    \Omega_t =  \left( \frac{||\frac{\partial \mathcal{E}}{\partial \textbf{h}_{t+1}} \frac{\partial \textbf{h}_{t+1}}{\partial \textbf{h}_{t}}||}{||\frac{\partial \mathcal{E}}{\partial \textbf{h}_{t+1}}||} - 1 \right)^2
\end{equation}

\noindent applied to the gradients of the transition matrix as follows:

\begin{equation}\label{eq:apply-omega}
\nabla \boldsymbol{W}^t_{hh} \gets \nabla \boldsymbol{W}^t_{hh} + \lambda \times \frac{\partial \Omega_t}{\partial  \boldsymbol{W}^t_{hh}}
\end{equation}

\noindent where $\lambda$ is a regularization coefficient. This regularization forces $\frac{\partial \textbf{h}_{t+1}}{\partial \textbf{h}_{t}}$ to preserve the norm in the correct direction determined by the error $\frac{\partial \mathcal{E}}{\partial \textbf{h}_{t+1}}$. Pascanu et al.~\cite{pmlr-v28-pascanu13} show that this regularization enables RNNs to handle longer sequences. Once calculated, $\Omega$ can also be reused to boost the gradients of the input mappings:

\begin{equation}
\nabla \boldsymbol{W}^t_{xh} \gets \nabla \boldsymbol{W}^t_{xh} + \lambda \times \frac{\partial \Omega_t}{\partial  \boldsymbol{W}^t_{xh}}
\end{equation}

The regularization suggested in \eqref{eq:omega} and \eqref{eq:apply-omega} is implemented in Algorithm~\ref{alg:regularisation}. The \texttt{Regularize} function also performs normalization. This is required as the regularization operation (line 4, Algorithm~\ref{alg:regularisation}) can lead to $||\nabla A_{reg}|| > 1$, fulfilling the necessary condition for the gradients to explode.

\begin{algorithm}
\caption{Regularization in the direction of update}\label{alg:regularisation}
\textbf{Input:} Local loss $\mathcal{L}(\textbf{z}_t, \hat{\textbf{z}}_t)$, hidden states $\textbf{h}_t$ and $\textbf{h}_{t+1}$, weight matrix $A$, gradient $\nabla A$, transition matrix $\textbf{W}_{hh}$, regularization coefficient $\lambda$, normalization constant $c$ \\
\textbf{Output:} Regularized gradient $\nabla A_{reg}$
    \begin{algorithmic}[1]
        \STATE \textbf{function} \texttt{Regularize} ($\mathcal{L}(\textbf{z}_t, \hat{\textbf{z}}_t)$, $\textbf{h}_t$, $\textbf{h}_{t+1}$, $A$, $\nabla A$, $\textbf{W}_{hh}$, $\lambda$, $c$) \label{lst:regularise}
            \vspace{1mm}
            \STATE \hspace{1em} $\nabla \mathcal{L}_{h_{t+1}} \gets \frac{\partial \mathcal{L}(\textbf{z}_t, \hat{\textbf{z}}_t)}{\partial h_{t+1}}$
            \STATE \hspace{1em} $\Omega \gets \frac{||\nabla \mathcal{L}_{h_{t+1}} \frac{\partial \textbf{h}_{t+1}}{\partial \textbf{h}_{t}}||}{||\nabla \mathcal{L}_{h_{t+1}}||} = \frac{||\nabla \mathcal{L}_{h_{t+1}} \textbf{W}_{hh} \text{diag} (\sigma'(\textbf{h}_t))||}{||\nabla \mathcal{L}_{h_{t+1}}||}$
            \item[]
            \STATE \hspace{1em} $\nabla A_{reg} \gets \nabla A + \lambda \times \frac{\partial \Omega}{\partial A}$ \label{lst:reg_op}
            \STATE \hspace{1em} $\nabla A_{reg} \gets \text{Normalize} (\nabla A_{reg}, c)$        
            \vspace{1mm}
            \STATE \hspace{1em} \textbf{return} $\nabla A_{reg}$    
    \end{algorithmic}
\end{algorithm}

In the context of \texttt{LRA-diff}, leveraging this regularization scheme is implemented by simply replacing the calls to \texttt{Normalize} with invocations of the \texttt{Regularize} function (lines 16-18 and line 21 in Algorithm~\ref{alg:lra-diff}). Note, that this has the ancillary effect of also boosting the local loss $\mathcal{L}(\textbf{z}_t, \hat{\textbf{z}}_t)$ as $\Delta \textbf{h}_{t-1}$ is also amplified.

\subsection{Computational demands}\label{sec:comp}
The dominant term in the overall complexity of the regularized version of Algorithm~\ref{alg:lra-diff} comes for the backward pass: $O(t_{max} \cdot K \cdot (|\boldsymbol{h}| \cdot |\boldsymbol{h}| + |\boldsymbol{x}| \cdot |\boldsymbol{h}|))$. Assuming that $\boldsymbol{h}$ and $\boldsymbol{x}$ are of similar magnitude, this simplifies to $O(t_{max} \cdot K \cdot |\boldsymbol{h}|^2)$. Given that the time complexity of TPTT is $O(t_{max} \cdot |\boldsymbol{h}|^2)$, it is clear that \texttt{LRA-diff} is worse by a factor of $K$. This prompts the question whether careful selection of $\gamma$ can eliminate the need to iterate over $K$, thus making the complexity of the two algorithms identical. 
\section{Experiments}\label{sec:Experiments}
In this section we present an empirical evaluation on the capability of \texttt{RNN-LRA-diff} to handle long sequences. We performed the evaluation using temporal order, 3-bit temporal order, and random permutation synthetic tasks, initially suggested in Hochreiter \& Schmidhuber~\cite{Hochreiter1997}.

Another set of experiments was performed to investigate the impact of using a different local loss. This was prompted by Ororbia et al.~\cite{ororbia2018conducting} who state that better results were achieved by using a log-penalty (Cauchy) loss:

\begin{equation}\label{eq:log-penalty}
    \mathcal{L}(\boldsymbol{z}, \boldsymbol{\hat{z}}) = \frac{1}{|\boldsymbol{z}|}\sum_{i=1}^{|\boldsymbol{z}|} \text{log} \left(1 + (z_i - \hat{z}_i)^2\right)
\end{equation}

For deeper investigation of the gradients flow, we also perform experiments that indiscriminately force the normalization of all gradients. This is implemented by omitting the $||\Delta|| \geq c$ check in \texttt{Normalize} (line 29 in Algorithm~\ref{alg:lra-diff}). The synthetic task experiments were also repeated with the modification suggested in Algorithm~\ref{alg:regularisation}, in order to evaluate the impact of the suggested regularization. 

\subsection{Implementation details}\label{sec:exp-details}
A simple RNN with hyperbolic tangent activations and learning rate $\alpha$ was trained using \texttt{RNN-LRA-diff} on the aforementioned tasks. The network was configured with 100 neurons in the transition matrix $\boldsymbol{W}_{hh}$. The size of the mapping matrices $\boldsymbol{W}_{xh}$ and $\boldsymbol{W}_{hy}$ was determined by the length of the input sequences and the number of target classes. The input symbols at each time step were one-hot encoded, and the initial weights were constructed using orthogonal initialization \cite{Henaff2016, Arjovsky2015}. The global loss $\mathcal{E}$ minimized by the network was set to the cross-entropy between the output $\hat{\boldsymbol{y}}$ and the target. Ororbia et al.~\cite{ororbia2018conducting} suggest several possible functions to track local loss. In our experiments, we use the mean squared error (MSE) and the log-penalty losses. The convergence criterion for considering a problem solved at a certain sequence length was taken to be a reduction of the global error to below $10^{-4}$, measured on a validation set of 10K samples. The training was terminated upon reaching the convergence criterion or after 100K iterations, whichever came first. A grid search was performed over $\alpha \in [10^{-1}, 10^{-2}, 10^{-3}]$, $\gamma \in [10^{1}, 10^{0}, 10^{-1}, 10^{-2}, 10^{-3}]$, and $K \in [10, 20, 30]$. The search space for
$\lambda$ was set to $\{2 \times 10^{0}, 10^{0}, 10^{-1}\}$. SGD~\citep{CAO20021, 10.1111/j.1467-985X.2006.00430_6.x} was used for applying the calculated weight updates. 

\subsection{Results}
\begin{figure*}[ht]
\begin{center}
\centerline{\includegraphics[width=0.9\textwidth]{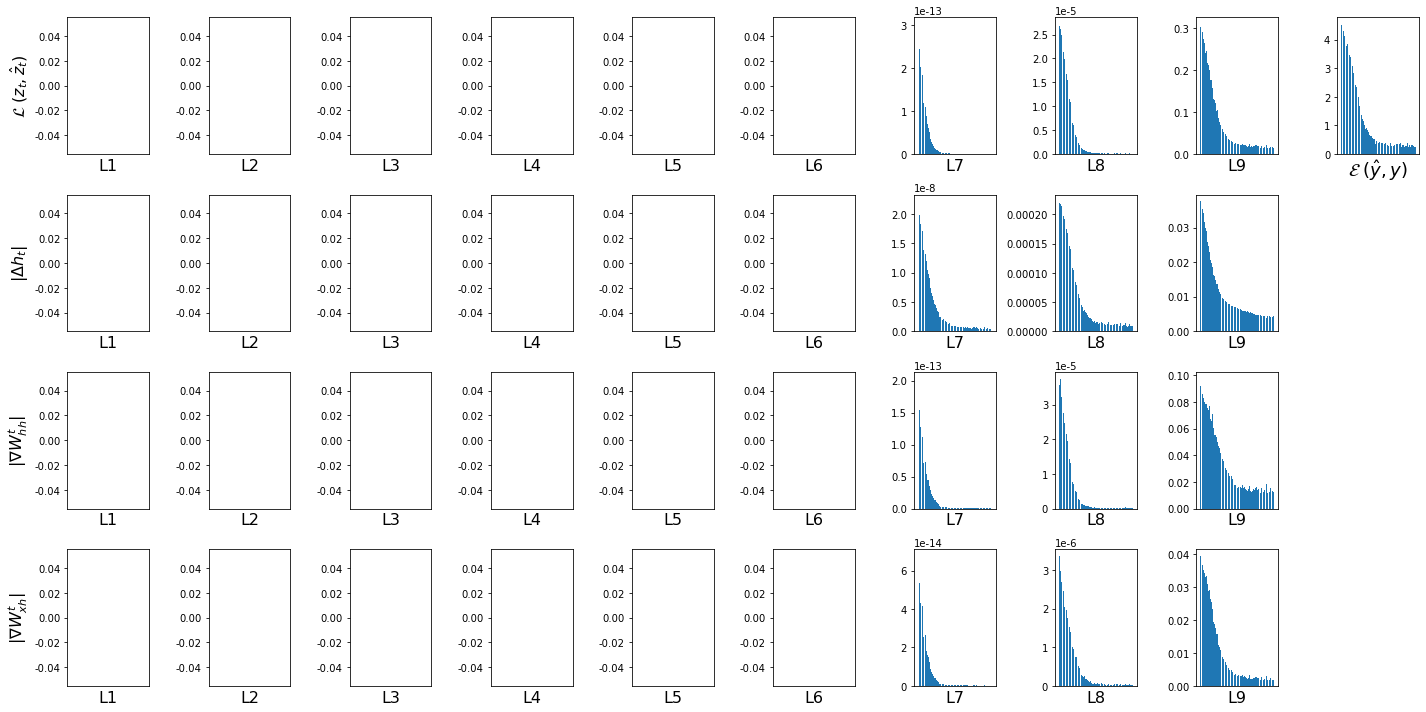}}
\caption{\textbf{Vanishing gradients in RNN-LRA-diff}. Results on the Random Permutation problem with sequence length $T=10$. Top row shows the change in losses over time for each layer (time step) of the network (i.e. $L1$ shows the change of $\mathcal{L}(\boldsymbol{z}, \boldsymbol{\hat{z}})$ over time for the first layer, $L2$ the change over time for the second layer and so on. The last subplot gives the change of the global loss $\mathcal{E}(\hat{y}, y)$). The second row shows the change in the norm of $\Delta \boldsymbol{h}_t$ for each layer of the network ($L1$ corresponds to time step $t_1$, $L2$ to $t_2$ and so on). The third row provides the same information for $|\nabla \boldsymbol{W}^t_{hh}|$. The bottom row gives the change over time of $|\nabla \boldsymbol{W}^t_{xh}|$. It is evident that the original algorithm demonstrates consistent reduction of the local loss, but its magnitude sharply decreases upstream --- the reduction of $\mathcal{L}(\boldsymbol{z}, \boldsymbol{\hat{z}})$ from L9 to L7 is twelve orders of magnitude, which is then cascaded down to the gradients computation.}
\label{fig:van-sub:1}
\end{center}
\end{figure*}

\begin{figure*}[ht]
\begin{center}
\centerline{\includegraphics[width=0.9\textwidth]{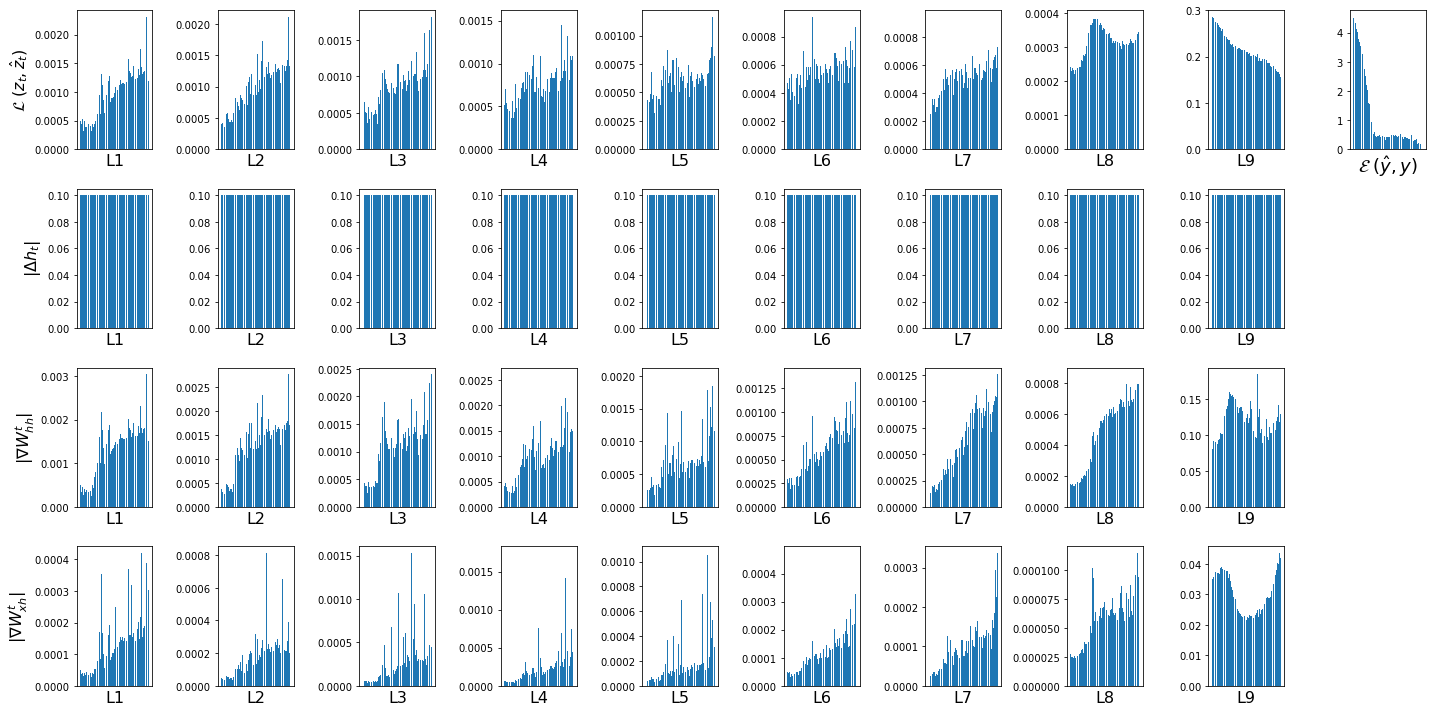}}
\caption{\textbf{Vanishing gradients in RNN-LRA-diff} --- Network tested on the Random Permutation problem with sequence length $T=10$. If normalization of $\Delta h_t$ via $c_1=1.0$ is always enforced, gradients flow through the entire network. The top row shows the change in losses over time for each layer of the network. The second row of plots shows the change in the norm of $\Delta \boldsymbol{h}_t$ for each layer of the network ($L1$ corresponds to time step $t_1$, $L2$ to $t_2$ and so on). The third row provides the same information for $|\nabla \boldsymbol{W}^t_{hh}|$. The bottom row gives the change over time of $|\nabla \boldsymbol{W}^t_{xh}|$ per layer.}
\label{fig:van-sub:2}
\end{center}
\end{figure*}

\begin{figure*}[t]
\begin{center}
\centerline{\includegraphics[width=0.9\textwidth]{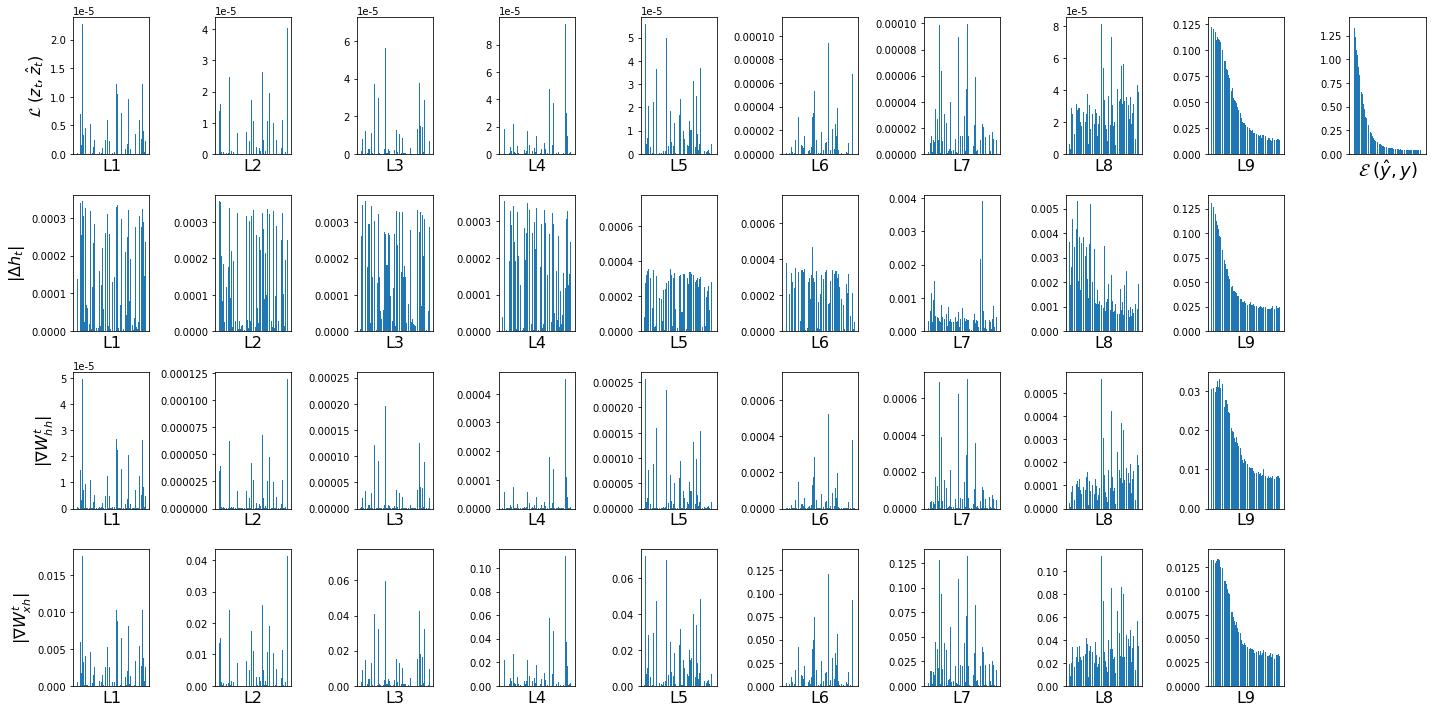}}
\caption{\textbf{Vanishing gradients in RNN-LRA-diff} --- Network tested on the Random Permutation problem with a sequence length $T=10$. $\Delta \boldsymbol{h}$, $\nabla \boldsymbol{W}_{hh}$, and $\nabla \boldsymbol{W}_{xh}$ are regularized using $\lambda=0.1$. The top row shows the change in losses over time for each layer (time step) of the network. The second row of plots shows the change in the norm of $\Delta \boldsymbol{h}_t$ for each layer of the network. The next two rows provide information on $|\nabla \boldsymbol{W}^t_{hh}|$ and $|\nabla \boldsymbol{W}^t_{xh}|$.}
\label{fig:van-sub:3}
\end{center}
\end{figure*}

Table~\ref{tab:rnn-lra-diff-synth} provides the results of Algorithm~\ref{alg:lra-diff} on the synthetic problems. Table~\ref{tab:rnn-lra-diff-synth-local-loss} shows the effect of using different local loss functions. Tables \ref{tab:rnn-lra-diff-synth-norm-impact} and \ref{tab:grid-all-norms} report results on the impact of normalization, and Table~\ref{tab:tptt-compare-reg} shows the model performance with the Algorithm~\ref{alg:regularisation} modification. Figures \ref{fig:van-sub:1}, \ref{fig:van-sub:2}, and \ref{fig:van-sub:3} show the gradient flow in the unmodified \texttt{RNN-LRA-diff}, \texttt{RNN-LRA-diff} with forced normalization, and regularization in the direction of the gradient, respectively.

\begin{table}
\begin{center}
\caption{\textbf{RNN-LRA-diff --- accuracy on the synthetic problems.} The \checkmark symbol denotes that the problem was solved in accordance  with the success criteria established in Section~\ref{sec:exp-details}.}
\label{tab:rnn-lra-diff-synth}
\begin{tabular}{lllll|lll|}
\textbf{Problem} &  \textbf{K} & \multicolumn{1}{|l|}{\textbf{T=10}} & \multicolumn{1}{l|}{\textbf{T=20}} & \textbf{T=30} \\
\hline
Temporal Order& 10  &  \multicolumn{1}{|c|}{\checkmark} & \multicolumn{1}{c|}{99.84} & \multicolumn{1}{c|}{84.02}  \\   
         & 20  & \multicolumn{1}{|c|}{\checkmark} & \multicolumn{1}{c|}{99.81} & \multicolumn{1}{c|}{84.01} \\   
         & 30  & \multicolumn{1}{|c|}{\checkmark} & \multicolumn{1}{c|}{99.84} & \multicolumn{1}{c|}{84.02}  \\            
\hline
3-bit Temporal Order& 10  & \multicolumn{1}{|c|}{\checkmark} & \multicolumn{1}{c|}{99.74} & \multicolumn{1}{c|}{71.09}  \\   
         & 20 & \multicolumn{1}{|c|}{\checkmark} & \multicolumn{1}{c|}{99.52} & \multicolumn{1}{c|}{70.60}  \\   
         & 30  & \multicolumn{1}{|c|}{\checkmark} & \multicolumn{1}{c|}{99.39} & \multicolumn{1}{c|}{68.42}  \\
\hline
Random & 10  &  \multicolumn{1}{|c|}{50.90} & \multicolumn{1}{c|}{51.06} & \multicolumn{1}{c|}{50.91} \\
permutation & 20  & \multicolumn{1}{|c|}{50.54} & \multicolumn{1}{c|}{51.07} & \multicolumn{1}{c|}{50.92}  \\
& 30  & \multicolumn{1}{|c|}{50.93} & \multicolumn{1}{c|}{51.23} & \multicolumn{1}{c|}{50.98} \\
\hline
\end{tabular} 
\end{center}
\end{table}

\begin{table*}[t]
\begin{center}
\caption{\textbf{RNN-LRA-diff --- Impact of local loss on the performance of the network.} Accuracy results from an identical grid search performed with a log-penalty local loss and MSE. All runs use normalization constants $c_0$ and $c_1$ set to 1.0.}
\label{tab:rnn-lra-diff-synth-local-loss}
\begin{tabular}{lllll|lll|lll}
        &   & \multicolumn{3}{c|}{\textbf{log-penalty}} & \multicolumn{3}{c}{\textbf{MSE}} \\
\textbf{Problem} & \textbf{K} & \multicolumn{1}{|l|}{\textbf{T=10}} & \multicolumn{1}{l|}{\textbf{T=20}} & \textbf{T=30} & \multicolumn{1}{l|}{\textbf{T=10}} & \multicolumn{1}{l|}{\textbf{T=20}} & \textbf{T=30} \\
\hline
Temporal Order& 10  & \multicolumn{1}{|c|}{\checkmark} & \multicolumn{1}{c|}{99.84} &  \multicolumn{1}{c|}{61.99} & \multicolumn{1}{c|}{\checkmark} & \multicolumn{1}{c|}{99.84} & \multicolumn{1}{c|}{83.50}  \\   
         & 20  & \multicolumn{1}{|c|}{\checkmark} & \multicolumn{1}{c|}{99.81} &  \multicolumn{1}{c|}{79.66} & \multicolumn{1}{c|}{\checkmark} & \multicolumn{1}{c|}{99.81} & \multicolumn{1}{c|}{82.32} \\   
         & 30  & \multicolumn{1}{|c|}{\checkmark} & \multicolumn{1}{c|}{99.84} &  \multicolumn{1}{c|}{79.35} & \multicolumn{1}{c|}{\checkmark} & \multicolumn{1}{c|}{99.84} & \multicolumn{1}{c|}{80.98}  \\            
\hline
3-bit Temporal Order& 10  & \multicolumn{1}{|c|}{\checkmark} & \multicolumn{1}{c|}{99.74} &  \multicolumn{1}{c|}{71.09} & \multicolumn{1}{c|}{\checkmark} & \multicolumn{1}{c|}{99.74} & \multicolumn{1}{c|}{71.09}  \\   
         & 20  & \multicolumn{1}{|c|}{\checkmark} & \multicolumn{1}{c|}{99.52} &  \multicolumn{1}{c|}{70.60} & \multicolumn{1}{c|}{\checkmark} & \multicolumn{1}{c|}{99.52} & \multicolumn{1}{c|}{70.60}  \\   
         & 30  & \multicolumn{1}{|c|}{\checkmark} & \multicolumn{1}{c|}{99.39} &  \multicolumn{1}{c|}{68.42} & \multicolumn{1}{c|}{\checkmark} & \multicolumn{1}{c|}{99.39} & \multicolumn{1}{c|}{68.42}  \\
\hline
Random & 10  & \multicolumn{1}{|c|}{50.54} & \multicolumn{1}{c|}{51.06} &  \multicolumn{1}{c|}{50.91} & \multicolumn{1}{c|}{50.90} & \multicolumn{1}{c|}{51.06} & \multicolumn{1}{c|}{50.91} \\
permutation & 20  & \multicolumn{1}{|c|}{50.54} & \multicolumn{1}{c|}{51.08} &  \multicolumn{1}{c|}{50.91} & \multicolumn{1}{c|}{50.54} & \multicolumn{1}{c|}{51.07} & \multicolumn{1}{c|}{50.92}  \\
& 30  & \multicolumn{1}{|c|}{50.93} & \multicolumn{1}{c|}{51.23} &  \multicolumn{1}{c|}{50.98} & \multicolumn{1}{c|}{50.93} & \multicolumn{1}{c|}{51.23} & \multicolumn{1}{c|}{50.98} \\
\hline
\end{tabular} 
\end{center}
\end{table*}

\begin{table*}[t]
\begin{center}
\caption{\textbf{RNN-LRA-diff --- Impact of normalization on the performance of the network}. The "No Normalization" section of the table presents accuracy results from an identical grid search run with the normalization constants $c_0$ and $c_1$ set to 0.}
\label{tab:rnn-lra-diff-synth-norm-impact}
\begin{tabular}{lllll|lll|lll}
        &   & \multicolumn{3}{c|}{\textbf{No Normalization}} & \multicolumn{3}{c}{\textbf{Normalization}} \\
\textbf{Problem} & \textbf{K} & \multicolumn{1}{|l|}{\textbf{T=10}} & \multicolumn{1}{l|}{\textbf{T=20}} & \textbf{T=30} & \multicolumn{1}{l|}{\textbf{T=10}} & \multicolumn{1}{l|}{\textbf{T=20}} & \textbf{T=30} \\
\hline
Temporal Order& 10  & \multicolumn{1}{|c|}{\checkmark} & \multicolumn{1}{c|}{99.84} &  \multicolumn{1}{c|}{83.13} & \multicolumn{1}{c|}{\checkmark} & \multicolumn{1}{c|}{99.84} & \multicolumn{1}{c|}{84.02}  \\   
         & 20  & \multicolumn{1}{|c|}{\checkmark} & \multicolumn{1}{c|}{99.81} &  \multicolumn{1}{c|}{81.45} & \multicolumn{1}{c|}{\checkmark} & \multicolumn{1}{c|}{99.81} & \multicolumn{1}{c|}{84.01} \\   
         & 30  & \multicolumn{1}{|c|}{\checkmark} & \multicolumn{1}{c|}{99.84} &  \multicolumn{1}{c|}{80.76} & \multicolumn{1}{c|}{\checkmark} & \multicolumn{1}{c|}{99.84} & \multicolumn{1}{c|}{84.02}  \\            
\hline
3-bit Temporal Order& 10  & \multicolumn{1}{|c|}{\checkmark} & \multicolumn{1}{c|}{99.74} &  \multicolumn{1}{c|}{70.80} & \multicolumn{1}{c|}{\checkmark} & \multicolumn{1}{c|}{99.74} & \multicolumn{1}{c|}{71.09}  \\   
         & 20  & \multicolumn{1}{|c|}{\checkmark} & \multicolumn{1}{c|}{99.52} &  \multicolumn{1}{c|}{69.44} & \multicolumn{1}{c|}{\checkmark} & \multicolumn{1}{c|}{99.52} & \multicolumn{1}{c|}{70.60}  \\   
         & 30  & \multicolumn{1}{|c|}{\checkmark} & \multicolumn{1}{c|}{99.39} &  \multicolumn{1}{c|}{66.88} & \multicolumn{1}{c|}{\checkmark} & \multicolumn{1}{c|}{99.39} & \multicolumn{1}{c|}{68.42}  \\
\hline
Random & 10  & \multicolumn{1}{|c|}{50.90} & \multicolumn{1}{c|}{51.06} &  \multicolumn{1}{c|}{50.91} & \multicolumn{1}{c|}{50.90} & \multicolumn{1}{c|}{51.06} & \multicolumn{1}{c|}{50.91} \\
permutation & 20  & \multicolumn{1}{|c|}{50.54} & \multicolumn{1}{c|}{51.07} &  \multicolumn{1}{c|}{50.92} & \multicolumn{1}{c|}{50.54} & \multicolumn{1}{c|}{51.07} & \multicolumn{1}{c|}{50.92}  \\
& 30  & \multicolumn{1}{|c|}{50.93} & \multicolumn{1}{c|}{51.23} &  \multicolumn{1}{c|}{50.98} & \multicolumn{1}{c|}{50.93} & \multicolumn{1}{c|}{51.23} & \multicolumn{1}{c|}{50.98} \\
\hline
\end{tabular} 
\end{center}
\end{table*}

\begin{table}
\begin{center}
\caption{\textbf{Impact of Normalization} --- Comprehensive grid search, including $c_0, c_1 \in \{0$, $10^{-2}$, $10^{-1}$, $10^{0}\}$ on the Temporal Order problem for $T=20$.}
\label{tab:grid-all-norms}
\begin{tabular}{lllll}
K & Learning rates & $c_0$ & $c_1$ & Accuracy \\ 
\hline 
10 & $\alpha=0.1$, $\gamma=0.01$ & 0 & 0.1 & 99.84 \\ 
20 & $\alpha=0.1$, $\gamma=0.01$ & 0 & 0.1 & 99.84 \\ 
30 & $\alpha=0.1$, $\gamma=0.01$ & 0 & 0.1 & 99.84 \\ 
\hline 
\end{tabular} 
\end{center}
\end{table}

\begin{table*}
\begin{center}
\caption{\textbf{Synthetic problems} --- Comparing TPTT and BPTT against RNN-LRA-diff and regularized RNA-LRA-diff using maximal sequence length achieved by model. Each model is trained with an initial sequence length of $T=10$. Upon successful convergence, $T$ is increased by 10 and the model is trained and tested again. The process is repeated until failure and the highest $T$ achieved by the network is reported. This protocol follows \cite{JMLR:v21:18-141} and the data in columns marked with $^{*}$ is directly sourced from there too.}
\label{tab:tptt-compare-reg}
\begin{tabular}{lcccc}
\textbf{Problem} & $\textbf{TPTT}^{*}$ & $\textbf{BPTT}^{*}$ & \textbf{LRA-RNN-diff} & \textbf{LRA-RNN-diff}\\ 
& & &  & \textbf{(regularized)}\\
\hline 
 Temporal Order& 150 & 120 & 10 & 20\\ 
3-bit Temporal Order & 150 &  70& 10& 10\\ 
Random Permutation & 300 & 70 &  -& 80\\ 
\hline 
\end{tabular} 
\end{center}
\end{table*}

\begin{figure*}[ht]
\begin{center}
\centerline{\includegraphics[width=0.95\textwidth]{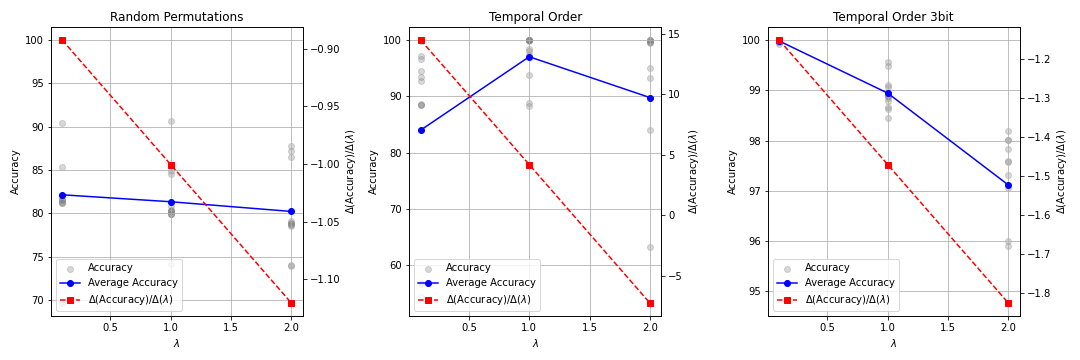}}
\caption{\textbf{Sensitivity of Regularization Coefficient $\lambda$} --- Impact of different values of $\lambda$ on the Random Permutations problem (T=20), Temporal Order problem (T=20), and 3-bit Temporal Order problem (T=10). Lower values of alpha generally lead to higher network accuracies across all three problem types.}
\label{fig:sens}
\end{center}
\end{figure*} 
\section{Discussion}\label{sec:discuss}

The results in Table~\ref{tab:rnn-lra-diff-synth} reveal that \texttt{RNN-LRA-diff} performs surprisingly poorly and in stark contrast to expectations and the original claims. The network manages to solve two of the problems only at a trivial depth of $T=10$, and it fails to solve the random permutations problem even for $T=10$. In contrast, TPTT can solve both temporal order problems for sequences of length of up to 150 time steps. TPTT can also successfully solve the random permutations problem for $T=300$. Even BPTT, which we established is extremely susceptible to gradient instabilities, can solve the in-scope problems for longer sequence lengths (see Table~\ref{tab:tptt-compare-reg}). We also observe that the value of $K$ has a marginal impact on the model performance (Table~\ref{tab:rnn-lra-diff-synth-local-loss} and Table~\ref{tab:rnn-lra-diff-synth-norm-impact}).

One possible explanation for the poor performance of \texttt{LRA-diff} could be that the MSE loss is not well suited for the algorithm. However, a log-penalty loss instead does not lead to noticeable improvement (see Table~\ref{tab:rnn-lra-diff-synth-local-loss}).

Another possibility for the poor performance is LRA's strong susceptibility to vanishing gradients. This is investigated in Figure~\ref{fig:van-sub:1}, which shows the change over time of key metrics such as the local loss $\mathcal{L}(\boldsymbol{z}, \boldsymbol{\hat{z}})$, the norm of $\Delta \boldsymbol{h}_t$, and the change in $|\nabla \boldsymbol{W}^t_{hh}|$. This information is presented for each layer of an \texttt{RNN-LRA-diff} network, unrolled over ten time-steps (L1 is the first layer of the network and L9 is the second to last layer. The final layer calculates the global loss $\mathcal{E}(\hat{y}, y)$ of the network). The data confirms that the global loss $\mathcal{E}$ is consistently reducing. However, inspecting $\mathcal{L}(\boldsymbol{z}_t, \boldsymbol{\hat{z}}_t)$ reveals that its magnitude sharply declines as we go back through the individual time steps. In fact, the local loss becomes zero only after three time steps, and no relevant information is being captured by the training process in any of the first six layers of the unfolded RNN. We can conclude that even if the network converges during training, the learning is shallow and mostly happens in the last three layers of the RNN. We also observe how the vanishing local loss directly impacts all other calculations, inhibiting updates (Figure~\ref{fig:van-sub:1}). The norms of $\nabla\boldsymbol{W}_{hh}$ and $\nabla\boldsymbol{W}_{xh}$ instantly go to zero as soon as the local loss vanishes for that particular layer. We can trace this issue to the original definition of the LRA algorithm. It is immediately apparent that the \texttt{Normalize} function in Algorithm~\ref{alg:lra-diff} activates only in the case of instabilities caused by exploding gradients. There are no provisions for handling vanishing gradients, and it appears that the claim that gradients do not vanish because training is performed locally does not hold. This is further supported by the results in Table~\ref{tab:rnn-lra-diff-synth-norm-impact}, where $c_0$ and $c_1$ are set to 0, effectively bypassing all normalization operations. We observe that normalization has no significant effect. Visual inspection of local losses and gradients also does not suggest an impact. Moreover, a comprehensive grid search that includes $c_0$ and $c_1$ also does not help to solve the Temporal Order for $T=20$ (Table~\ref{tab:grid-all-norms}), further indicating that the normalization values do not play a significant role.

One na\"ive fix for the vanishing gradients is to indiscriminately amplify their norm, which will in turn drive the magnitude of the local losses. This type of forced normalization is a useful test to confirm that the architecture is indeed susceptible to vanishing gradients. The results in Figure~\ref{fig:van-sub:2} confirm that when indiscriminate enforcement takes place, gradients and errors do flow through all layers. Unfortunately, despite the improved gradient flow no significant improvement in learning is observed. This type of artificial amplification of $||\frac{\partial \boldsymbol{h}_{t_{max}}}{\partial \boldsymbol{h}_t}||$ makes $\mathcal{E}$ more sensitive to all inputs $(x_1, x_2, \dots, x_{t_{max}})$. Despite being a useful test, this hard constraint hurts training as the network cannot learn to ignore certain inputs in order to successfully solve the problem.

The regularized version of \texttt{LRA-RNN-diff} outperforms Algorithm~\ref{alg:lra-diff}. It can successfully solve the Temporal Problem at twice the original depth. Moreover, after the regularization modification, the previously unsolvable Random Permutation problem is now successfully solved at depths comparable to those achieved by BPTT. It is worth noting that \texttt{LRA-RNN-diff} also achieves better accuracy for $T=20$ on the 3-bit Temporal Order problem, however, this improvement is not sufficient to match the established success criterion; therefore, we still consider the regularized version as identically performing to the vanilla one. In addition, we observe that in the regularized \texttt{LRA-RNN-diff}, the magnitude of $\Delta \boldsymbol{h}_t$ varies and gradients keep flowing all the way to the top upstream layers. This is in contrast with the shallow learning of the original LRA algorithm and further reinforces the validity of the hypothesis that vanishing gradients exert a negative influence on LRA learning. These results confirm that LRA does indeed suffer from vanishing gradients, and the claim in \cite{ororbia2018conducting} that such effects are avoided by treating the network as a series of locally optimized sub-graphs does not hold. It should be emphasized that no claim is made that regularization in the direction of the update resolves the deficiencies of \texttt{LRA-diff} in handling long-term dependencies. Rather, it is employed as a tool to further validate that \texttt{LRA-diff}’s limitations in this regard are attributable to vanishing gradients.

We also looked at the sensitivity of the network with regard to the regularization parameter $\lambda$. Figure~\ref{fig:sens} shows the relative change $\Delta (Accuracy) / \Delta (\lambda)$ using results from the grid search on the Random Permutations (T=20), Temporal Order (T=20), and 3-bit Temporal Order (T=10) problems. We observe that lower values of $\lambda$ generally lead to higher improvements in accuracy. This is in contrast with the 2.0 fixed value used in \cite{pmlr-v28-pascanu13},  but is expected as the LRA is substantially different from backpropagation. It is also important to note that the optimal alpha value may depend on other hyperparameters, such as $c$ and $K$. 
\section{Conclusion and future work}

In this paper, we proposed and assessed the suitability of Local Representation Alignment for training RNNs. We conclude that LRA can train RNNs to a certain task-specific depth, but still under-performs in comparison to state of the art target propagation architectures like TPTT. We empirically established that this is due to LRA's susceptibility to vanishing gradients -- a problem that can be mitigated by using regularization in the direction of the update. 

It should be acknowledged that our study only looked at the \texttt{LRA-diff} algorithm as a potential training method. The deficiencies found in the suitability of training RNNs with LRA are therefore strictly constrained to the \texttt{LRA-diff} variant of the algorithm. The applicability of these limitations to all types of LRA algorithms remains an open question, warranting further investigation into other LRA variants, such as \texttt{LRA-fdbk}, \texttt{LRA-E}, and \texttt{Rec-LRA}~\cite{ororbia2023backpropagation}. Future work will be necessary to determine the extent to which these findings generalize.

It should be noted that the regularized \texttt{RNN-LRA-diff} relies on a fixed coefficient $\lambda$. This could lead to suboptimal performance, and we envision a modification where $\lambda$ is dynamically adjusted on a layer-by-layer basis and driven by the magnitude of the gradient of the neighboring downstream layer. We leave this as an area for further exploration.

During our experiments, optimizing the number of iterations $K$ over the pre-activation displacement showed to confer minimal advantage. It appears that keeping the number of iterations to a minimum or even eliminating the inner loop altogether can be fully compensated by using a fine-tuned value for the target adjustment step $\gamma$. This could not only yield improved results, but can substantially reduce the computational demands of the algorithm. However, a more thorough analysis of the inner-loop behavior is needed to validate this assumption.

Furthermore, the performance evaluation was purely based on synthetic problems. Similar RNN evaluations typically include problems based on real-world datasets. Two such examples are the MNIST Sequence of Pixels task \cite{44961} and the Polyphonic Music task \cite{https://doi.org/10.1111/coin.12691}. Unfortunately, both tasks require the evaluated RNN to handle sequences of greater length ($T=784$ and $T=100$, respectively). Therefore, we must defer this evaluation until future improvements to \texttt{RNN-LRA-diff} make it possible.

\bibliography{main}
\bibliographystyle{IEEEtran}

\end{document}